\documentclass[sigconf]{acmart}

\usepackage{pgfplots}
\usepackage{algpseudocode}
\usepackage{graphicx}
\usepackage{latexsym}
\usepackage{algorithm}
\usepackage{fix-cm}

\AtBeginDocument{%
  }

\setcopyright{acmlicensed}
\copyrightyear{2025}
\acmYear{2025}
\acmDOI{XXXXXXX.XXXXXXX}
\acmConference[CODS '25]{Make sure to enter the correct
  conference title from your rights confirmation email}{December 17--20,
  2025}{Pune, India}
\acmISBN{978-1-4503-XXXX-X/2018/06}

\begin{document}

\title{Hierarchical Sparse Plus Low Rank Compression of LLM}


\author{Pawan Kumar}
\email{pawan.kumar@iiit.ac.in}
\orcid{1234-5678-9012}
\affiliation{%
  \institution{International Institute of Information Technology, Hyderabad}
  \city{Hyderabad}
  \state{Telangana}
  \country{India}
}

\author{Aditi Gupta}
\email{aditi.gu@research.iiit.ac.in}
\affiliation{%
  \institution{International Institute of Information Technology, Hyderabad}
  \city{Hyderabad}
  \state{Telangana}
  \country{India}
}

\renewcommand{\shortauthors}{Kumar et al.}


\begin{abstract}
Modern large language models (LLMs) place extraordinary pressure on memory and compute budgets, making principled compression indispensable for both deployment and continued training.  We present Hierarchical Sparse Plus Low-Rank (HSS) compression, a two–stage scheme that (i) removes the largest‐magnitude weights into a sparse matrix $S$ and (ii) applies a recursive Hierarchically Sparse Separable (HSS) low-rank factorisation to the dense residual matrix.  A recursive rank-reducing strategy and a reverse Cuthill–Mckee (RCM) permutation are introduced to align high weights towards the diagonal with the block-diagonal hierarchy, maximising off-diagonal compressibility (because they are touched only once).  HSS is hardware-friendly: its matrix–vector multiply reduces to one sparse and a sequence of thin-matrix multiplications, and can be trained end-to-end with standard optimisers.

Experiments on LLaMA-7B show that targeting only the self-attention projections (1.6B parameters of $ Q$, $ K$, and $ V$ matrices out of a total 7B parameters) suffices to yield large memory savings while retaining comparable state-of-the-art perplexity scores on test samples of the WikiText dataset.  For example, with a 30\% sparsity budget and an outer rank of 512, sHSS-RCM achieves a perplexity of 1.64, outperforming dense baselines and classical sparse-plus-SVD variants, while also achieving significant memory savings. \\ 
Code: {https://github.com/misterpawan/hi-solo-llm} 
\end{abstract}




\maketitle

\section{Introduction and Related Work}


When the Transformer model was first introduced in 2017 \cite{Vaswani2017Attention}, it laid the foundation for long-range dependency modeling in NLP, and catapulted to the top of the field driving state-of-the-art results in translation, summarization, QA, and elementary maths and science problems \cite{Devlin2019BERT,Mandlecha2022HybridTokenization,Chatakonda2021}. As the scale in parameters, data, and computing power increased, large language models (LLMs) showed strong in-context learning and emergence abilities that have been refined with instruction tuning to create aligned, general-purpose assistants  \cite{Brown2020LanguageModels,Ouyang2022InstructGPT}. Empirical predictions about predictable scaling and optimal training make it a mystery why smaller models don't quite keep up, no matter how hard they try  \cite{Kaplan2020ScalingLaws,Hoffmann2022Chinchilla}, but such advances in LLMs have now been spreading to coding, and maths and science, courtesy of  \cite{Lewkowycz2022Minerva,Chen2021Codex}. But, deploying these huge models is costing us in terms of memory, delay and energy consumption. Prompting researchers to cut down on the size of these models, by means of model compression, distillation \cite{Sanh2019DistilBERT}, pruning, low-rank adjustments and quantisation.


With respect to the growing scale of machine learning models, model compression \cite{zhu2024surveymodelcompressionlarge} is key to reducing the size, computational cost and memory footprint of those models, and not necessarily at the expense of their performance. For large language models and diffusion models, which, with their large number of parameters, are making efficient deployment, fine-tuning and running inference is a real challenge. The original GPT-3 model \cite{Brown2020LanguageModels} had 175 billion parameters, and recent diffusion models like Imagen \cite{saharia2022photorealistictexttoimagediffusionmodels} and Stable Diffusion \cite{rombach2022highresolutionimagesynthesislatent} require massive amounts of storage and GPU power for picture synthesis. Well-known for going beyond just tweaking for better efficiency, model compression is  about making cutting-edge AI accessible to everyone.


With respect to large language models (LLMs) and diffusion models, compression plays a critical role in the real-world deployment of these models. For latency-sensitive applications like chatbots and real-time image generators, that needs to be run on low-powered devices like mobile phones or edge devices for privacy reasons, makes these models impractical to run outside of massive datacentres. Well-known techniques, such as model compression, get around this problem by reducing the size of the model, and also make it possible to train, fine-tune, and run these models much more quickly, all the while using less energy. It also paves the way for completely private personalization through efficient on-device adaptation. Hugging Face’s DistilBERT \cite{sanh2020distilbertdistilledversionbert} and Meta’s LLaMA model \cite{touvron2023llamaopenefficientfoundation} families are examples of the results of significant compression efforts that didn’t sacrifice much in the way of accuracy. This process is basically a necessary step towards making AI more accessible to everyone, by removing the infrastructure barrier.


Researchers have turned to two complementary approaches: bit-level compression and rank-based compression, when compressing machine learning models. Bit-level compression \cite{gong2024surveylowbitlargelanguage} cuts down the precision of model parameters. Transforming them into 8-bit, 4-bit, or even 1-bit \cite{wang2023bitnet} representations, with the help of techniques such as quantization and integer-only inference, as shown in Wang et al. BitNet \cite{wang2023bitnet}. This method can drastically reduce model size and accelerate matrix operations, and was demonstrated in GPTQ \cite{frantar2023gptqaccurateposttrainingquantization}. Rank-based compression, on the other hand, involves reconstructing the weight matrices of a model by using lower rank structures, a concept pioneered by LoRA \cite{hu2021loralowrankadaptationlarge}, in which low-rank adapters are appended to pre-trained models, and/or through tensor decomposition and matrix factorization that disassemble dense matrices into a manageable handful of components. Sparse methods and pruning strategies, like Han et al.’s Deep Compression \cite{Han2016DeepCompression}, Li et al.’s Pruning Filters \cite{Li2017PruningFilters}, Mostafa et al.’s Dynamic Sparse \cite{Mostafa2019DynamicSparse} and Sanh et al.’s Movement Pruning \cite{Sanh2020MovementPruning}, can also be viewed as being part of the rank-related compression framework.

\begin{figure*}
    \centering
    \includegraphics[width=1.0\linewidth]{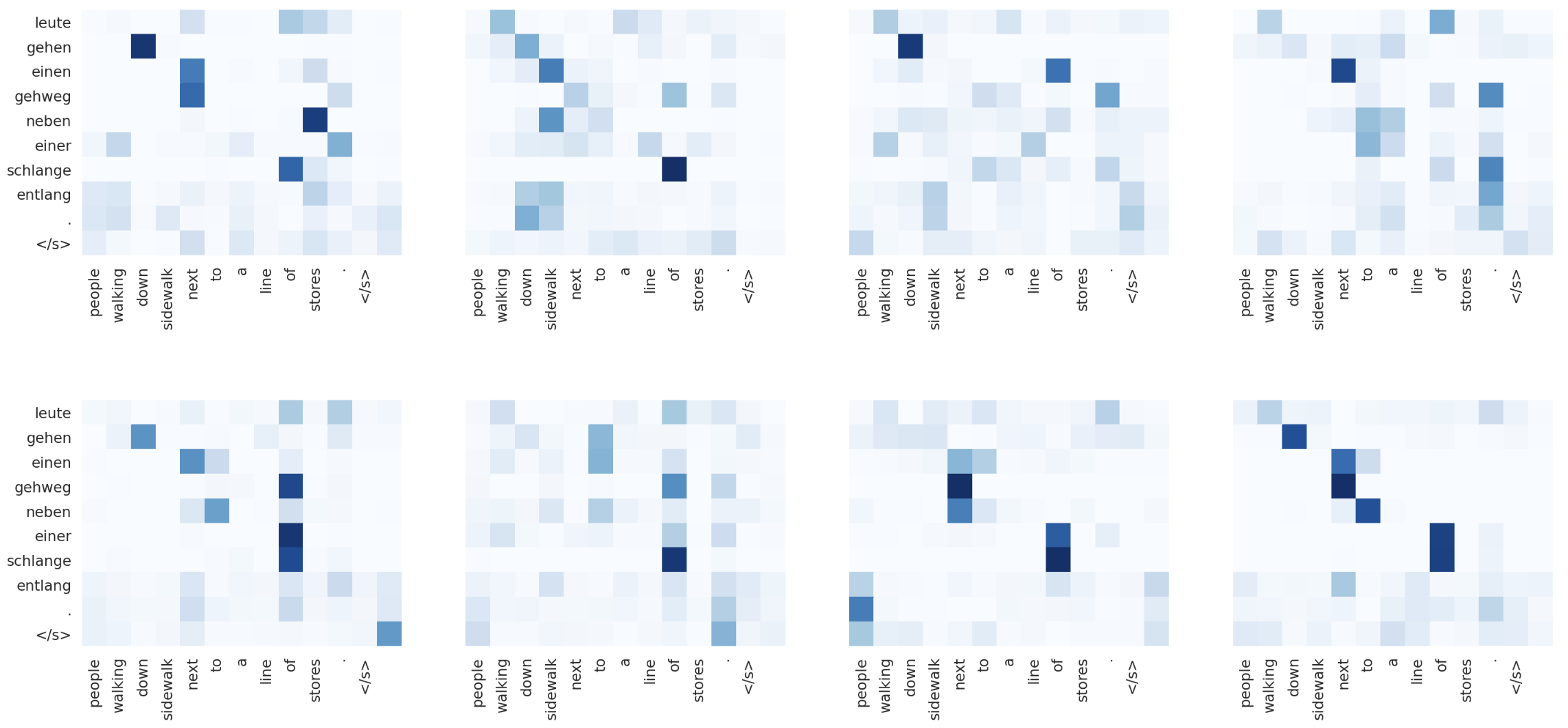}
    \caption{Plot of low-rankness in off-diagonal blocks of attention matrix for German-English translation. We can see a weaker correlation of attention to far away tokens (see as lesser attention scores in the attention matrix). Not to say that all far away tokens are not relevant, but often only a few are relevant, hence, indicating an opportunity for possible low rank compression of off-diagonal blocks. This figure was generated using the publicly available Google Colab notebook code \cite{gala2021attention}.}
    \label{fig:attention-matrix}
\end{figure*}


As for model compression, there's still a lot of unexplored territory, one of the biggest being the accuracy. Coming rushing into model compression by training a low-rank model or quantizing a model can lead to a nasty drop in performance. Well-known researchers have tackled this problem and improved methods for accurately quantizing a model in low-rank/low-bitwidth situations \cite{awq23,gptq22,smoothquant22,kvquant24,quantreason25,azizi24,ji24}. 
Another problem with compression is that different parts of the architecture can be affected differently. 
Some might not be very sensitive, but others might be, so one may need to allocate specific parameters for those weights or even use a combination of different precision for efficiency. 
We also can’t really say much about the training of models that have been compressed, take Large Language Models as an example: the convergence properties of training low-rank or quantized LLMs, from scratch, is still a mystery. Lastly, even in cases where compression is relatively straightforward, there’s the problem of scaling. Consider, for example, LLaMA 65B or U-Net, the backbone network for Diffusion.


With respect to model compression, one of the main concerns is the potential security vulnerability, especially in compressed models, due to the reduced representational capacity, which is a well-known issue in neural network research. Ye, Wang, Gorsline, Savostianova, Jian, Chu, and Piras in their respective works, showed in \cite{Ye2019,Wang2018,Gorsline2021,Savostianova2023,Jian2022,Chu2023,Piras2025} the potential attacks on compressed models. One recent development in model compression is the exploration of a combination of multiple techniques, i.e., sparsity, quantization and low-rank decomposition by hybrid approaches seen in SparseGPT \cite{frantar2023sparsegptmassivelanguagemodels} and QLoRA \cite{dettmers2023qloraefficientfinetuningquantized}. There are research that zeroes in on methods that are task-specific, hardware-sensitive, and theoretically grounded. As large-scale models become the heart of today's AI systems, model compression will not just be a necessity but will be at the forefront of what makes AI responsible and scalable.

In this work, we contribute towards compression algorithms by proposing a sparse plus {\em hierarchical} low rank compression method. The highlights of our work are as follows:

\begin{itemize}

\item Sparse + Hierarchical Low-Rank Compression.
We propose sHSS, a multilevel Hierarchically separable factorization that first carves out the top-$p$ \% high-magnitude weights into a sparse mask and then applies depth-halved low-rank approximations to the off-diagonal blocks of the residual matrix.
    
\item RCM-enhanced Variant.
By re-ordering weight matrices with Reverse Cuthill–Mckee (sHSS-RCM), off-diagonal ranks may shrink further, yielding lower perplexity than vanilla sHSS.  We consider sparse + HSS, sparse + exact SVD, and sparse + randomised SVD under one formulation of sparse plus (hierarchical) low rank and show a comparative study of these various compression techniques on Llama-7B weights; we investigate the role of sparsity, matrix reordering, and low rank compressions on perplexity scores.
    
\item Our results show that an end-to-end half-precision (16-bit) CUDA implementation compresses 1.6B parameters of linear layers of Llama-7B on a single H100, achieving up to $1.7 \times$ storage reduction with sometimes even better or on par perplexity compared to the uncompressed model and outperforms existing compression baselines.

\end{itemize}

The rest of the paper is organised as follows. In section 2, we discuss the motivation for hierarchical matrices. In section 3, we discuss SVD-based compression and sparse plus low rank approaches proposed recently \cite{han2024sltrainsparsepluslowrank} for LLM compression. In section 4, we introduce our HSS and sparse plus low rank HSS-based compression method. Finally, in section 5, we show numerical experiments on the tradeoff between compression and perplexity scores for various compression methods.

\section{Motivation for Hierarchical Low Rank}


Hierarchical low-rank approximation has proven to be a time-tested technique, when large-scale linear algebra operations are to be accelerated in the world of scientific computing. The systematic study and development of Hierarchical $\mathcal{H}$-matrix arithmetic and factorization methods were done by Hackbusch, Borm, and others. The field has seen the creation of a group of algorithms that are able to find near-linear solutions to dense linear systems, and are used in conjunction with preconditioned sparse linear solvers \cite{Larin2008ACS, Kressner2015PreconditionedLR, Bai2005StructuredPF, Stoll2008CombinationPA, Axelsson1985ASO, kumar2014b, kumar2013, Katyan2020TwoGrid, das2021}. Further low-rank or hierarchical low rank approximate preconditioners, were introduced \cite{Li2013DivideAC, Pouransari2015FastHS, Andreev2015MultilevelPA, Grasedyck2007DomaindecompositionB, Knyazev2007BlockLO, Bonev2022, parvizi2023hierarchicallupreconditioningtimeharmonic, szeliski2006}.


The Fast Multipole Method, or FMM, has been at the heart of many algorithms. It was proposed as a fast solver for reducing the complexity of $N$-body problems to $O(N)$ or $O(N log N).$ The algebraic equivalent of the FMM has spawned the development of hierarchical matrix formats including $H, H^2,$ HSS (Hierarchically Semi-Separable) and HODLR \cite{Hartland2023,Khan2024}. The basic premise of these formats is to take advantage of the fact that the off-diagonal blocks in matrices generated by discretised integral equations contain a lot of long-range interactions, and this property can be leveraged because these long-range interactions tend to be numerically low-rank, something that can be effectively captured by low-rank factorisations. 


When it comes to reducing the complexity of matrix multiplication in modern machine learning, a lot of significance has been given to large language models, with the use of attention and multi-layer perceptron layers leading to very large matrices that have inherent structure to them. The off-diagonal elements of the attention matrix, seen in $QK^T$, mirror the weak, often far-off connections between tokens in sequence space, and these could possibly be represented by low-rank approximations. Consider the example of an English-German translator in Figure \ref{fig:attention-matrix}. Here, we exploit the low rankness of matrices $W^Q, W^K, W^V$ for which $Q = XW^Q, K = XW^K, V = XW^V$. We should note that other papers have previously exploited the low-rank property of the attention matrix $QK^T$ \cite{wang2020linformerselfattentionlinearcomplexity,choromanski2022rethinkingattentionperformers}. But we focus on compressing $W^Q, W^K, W^V$ directly. Restricting the compression to only $QK^T$ will restrict the amount of compression we can achieve.

We leverage these insights to motivate our use of hierarchical sparse-plus-low-rank decompositions, which not only reduce parameter count and memory footprint but also mirror well-established techniques from high-performance numerical linear algebra. The resulting representations preserve key algorithmic advantages such as matrix–vector product efficiency, structured sparsity, and favourable scaling behaviour.


Researchers have started to use techniques such as block diagonalization, sparsity and hierarchical low-rank factorization, when applying deep learning techniques to transformers \cite{Tay2020_BlockLM,Dao2022_HTransformer1D}. For transformers, the technique we presented in this work is the first to take a closer look at the impact of matrix reordering, sparsity and hierarchical low-rank factorization in contemporary large language models, or LLMs. 

Our interest in sparse reordering arises from the fact that unlike in scientific computing, where diagonal blocks are theoretically full rank and off-diagonal blocks are low rank, this isn’t necessarily true in the case of projection weight matrices $W^Q, W^K, W^V$ in LLM layers. Reordering provides a slightly more effective compression in our results.

\section{Low Rank Compression Methods}

\section*{Low-Rank and Sparse Compression Methods}

Let $W\in\mathbb{R}^{m\times n}$ be a dense weight matrix drawn from a neural network layer.  
We study four compression strategies that preserve the predictive power of $W$ while reducing its memory footprint and compute cost.  
Each approach exploits either low-rank structure, explicit sparsity, or a combination of the two.

\subsection*{Truncated (Exact) SVD Compression}

The first method, is a well-known method, relies on the classical singular-value decomposition.  
By retaining only the top $k$ singular triplets, one obtains the rank-$k$ approximation
\[
  W \approx U_k\Sigma_kV_k^{\!\top},
\]
where $U_k\in\mathbb{R}^{m\times k}$ and $V_k\in\mathbb{R}^{n\times k}$ are orthonormal, and $\Sigma_k$ is diagonal.  
Because the truncated SVD is the optimal Frobenius-norm approximation of rank $k$, this method offers the best accuracy per parameter among linear factorizations. There are other ways of rank compression using QR factorizations \cite{golub2013matrix,kumar2014,kumar2015}. 
Storage drops from $mn$ parameters to $(m+n)k$; computationally, however, forming the SVD costs $O(mn\min\{m,n\})$, which is prohibitive for very large layers.  
In practice, the diagonal matrix can be absorbed into one factor, for instance, by redefining $U_k \leftarrow U_k\Sigma_k^{1/2}$ and $V_k^\top\leftarrow\Sigma_k^{1/2}V_k^\top$ so that inference reduces to two narrow matrix multiplications.

\begin{algorithm*}[t!]          
\caption{\label{alg:three-level}Three–Level HSS Compression with reordering. This algorithm does HSS compression for three levels. The input $A$ is the weight matrix to be compressed using HSS, tolerance $\epsilon$ can be used to consider singular values to retain. Also, parameter $k$ is used to keep the $k$ largest number of singular values. A sparse plus Hierarchical low rank version of the algorithm will take out some sparse components from $A, D_0, D_1$ before calling the {two\_level\_compress} method. This is explained in detail in Section \ref{sec:shss}.}
\begin{algorithmic}[1]
  \State \textbf{Input:} $A \in \mathbb{R}^{N\times N}$ to be compressed \State \textbf{Input:} $\varepsilon:$ tolerance for singular values
  \State \textbf{Input:} $k:$ number of largest singular values to retain
  \State \textbf{Top level:}\;
         $(D_{0},D_{1},U_{t0},R_{t0},U_{t1},R_{t1})
         \gets \texttt{two\_level\_compress}(A,\varepsilon,k)$ \Comment{apply RCM to $A$}
  \State \textbf{Left subtree:}\;
         $(D_{00},D_{01},U_{00},R_{00},U_{01},R_{01})
         \gets \texttt{two\_level\_compress}(D_{0},\varepsilon,k)$ \Comment{apply RCM to $D_0$}
  \State \textbf{Right subtree:}\;
         $(D_{10},D_{11},U_{10},R_{10},U_{11},R_{11})
         \gets \texttt{two\_level\_compress}(D_{1},\varepsilon,k)$ \Comment{apply RCM to $D_1$}
  \State \textbf{Return:}\;
         $\{D_{ij},U_{ij},R_{ij}\}_{i,j\in\{0,1\}}$
         and top–level $\bigl(U_{t\bullet},R_{t\bullet}\bigr)$
\end{algorithmic}
\end{algorithm*}

\subsection*{Randomized SVD Compression}

Randomized SVD replaces the expensive exact factorization with a ``sketch-based" approach.  
A Gaussian test matrix $\Omega\in\mathbb{R}^{n\times\ell}$ ($\ell\ge k$) is first drawn, and the product $Y=W\Omega$ is orthonormally decomposed as $Y=QR$.  
Projecting $W$ onto the subspace spanned by $Q$ yields the much smaller matrix $B=Q^{\!\top}W\in\mathbb{R}^{\ell\times n}$, on which an exact SVD is feasible.  
Setting $\widehat U_k = Q\tilde U_k$ recovers the approximate factors
$
  W \approx \widehat U_k\widehat\Sigma_k\widehat V_k^{\!\top}.
$
When $k\ll\min\{m,n\}$, the total cost falls to $O(mn\log k + (m+n)k^2)$ and can be streamed.  
Accuracy is slightly lower than the deterministic SVD but can be tightened by oversampling ($\ell=k+q$) and by one or two power iterations that amplify the singular spectrum. The high computation cost of SVD leads to a faster but approximate randomized SVD variant.

\subsection*{Sparse-Plus-Low-Rank: Exact SVD on the Residual}

The early work of Chandrasekaran et al \cite{CHANDRASEKARAN20091493,bertsimas2023sparsepluslowrank} popularised the sparse plus low-rank decomposition, which has its roots in model and system identification, a problem that's  NP-hard. Well-known weight matrices of modern large language models show a pattern of a few very large spikes and some relatively low-rank blocks, often observed in the attention matrix $QK^T.$

To exploit this property, we first take out a possibly sparse matrix
\[
  S \;=\; \operatorname{top}_{p\%}\!\bigl(|W|\bigr),
\]
which retains the largest \(p\)-percentile magnitudes of \(|W|\), and then form the residual
\(
  R = W - S.
\)
After these significant entries are removed, \(R\) is much expected to be more compressible and admits a rank-\(k\) SVD, so that the overall approximation becomes
\[
  W \;\approx\; S \;+\; U_k \Sigma_k V_k^{\!\top}.
\]
In an interesting recent paper SLTrain \cite{han2024sltrainsparsepluslowrank}, the LLM was trained assuming the weight matrices of linear layers of LLM to be in sparse plus low rank form. 
This hybrid representation (sparse plus low-rank) apparently preserves salient outliers exactly while still enjoying the storage savings of a compact low-rank factorization for the smoother component.  
Its chief drawback is computational: selecting the top entries requires sorting \(mn\) magnitudes: an \(O\!\bigl(mn\log(mn)\bigr)\) operation, followed by an SVD of the residual \(R\).

\subsection*{Sparse-Plus-Low-Rank: Randomized SVD on the Residual}

A more scalable variant applies the sparse mask but replaces the deterministic SVD on $R$ with the randomised procedure described earlier.  
Consequently,
\[
  W \approx S + \widehat U_k\widehat\Sigma_k\widehat V_k^{\!\top}.
\]
With often the same storage profile as the previous method, but markedly lower runtime especially when $k$ is small.  
Although the approximation error is marginally higher than using an exact SVD on the residual, empirical studies show that a modest oversampling budget compensates for most of the loss while keeping the overall pipeline tractable for billion-parameter models.

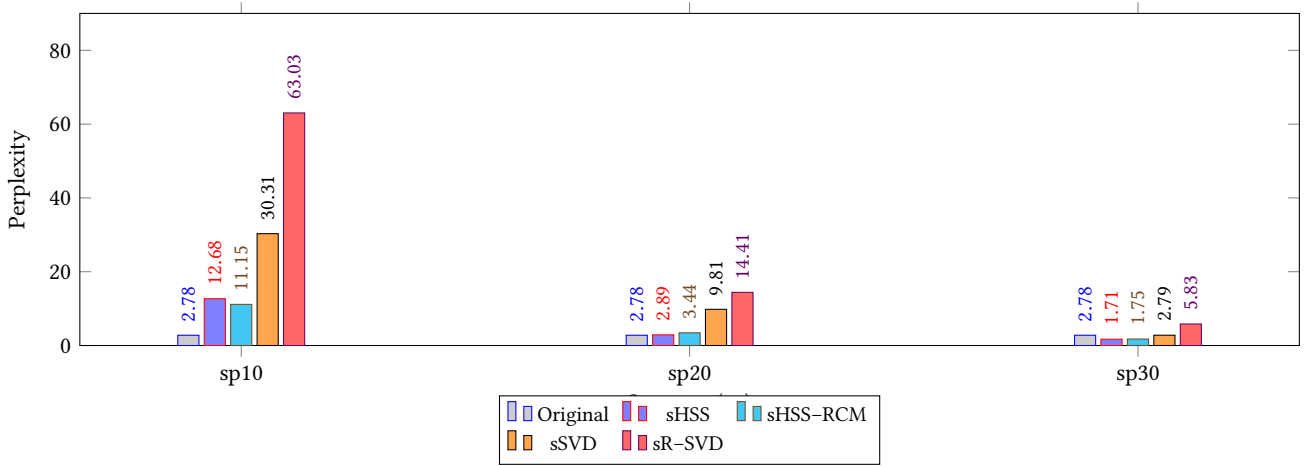
\begin{figure*}[t!]
\begin{tikzpicture}[scale=1.0] 
\begin{axis}[
    ybar,
    bar width=8pt,
    enlarge x limits=0.18,
    ymin=0,
    ymax=90,                    
    ylabel={Perplexity},
    ylabel style={yshift=-12pt},
    symbolic x coords={sp10, sp20, sp30},
    xtick=data,
    xlabel={Sparse\,\%  ($\text{sp}$)},
    legend columns=3,
    legend style={
        font=\small,
        /tikz/every even column/.append style={column sep=3pt},
        at={(0.5,-0.15)}, anchor=north},
    nodes near coords,
    nodes near coords style={
        yshift=2pt,
        font=\small,
        rotate=90,
        anchor=west,
        /pgf/number format/.cd, fixed,precision=2},
    width=\linewidth,
    height=6cm
]
\addplot+[fill=gray!40]  coordinates {(sp10,  2.78) (sp20,  2.78) (sp30, 2.78)};
\addplot+[fill=blue!50]  coordinates {(sp10, 12.68) (sp20,  2.89) (sp30, 1.71)};
\addplot+[fill=cyan!60]  coordinates {(sp10, 11.15) (sp20,  3.44) (sp30, 1.75)};
\addplot+[fill=orange!70]coordinates {(sp10, 30.31) (sp20,  9.81) (sp30, 2.79)};
\addplot+[fill=red!60]   coordinates {(sp10, 63.03) (sp20, 14.41) (sp30, 5.83)};
\legend{Original, sHSS, sHSS--RCM, sSVD, sR--SVD}
\end{axis}
\end{tikzpicture}
\caption{Ablation study for rank 512, depth = 4 for Hierarchical methods sHSS, sHSS-RCM. And sp10, sp20, sp30 show the percentage of the largest entries of weight stored in the separate matrix $S.$ \label{fig: ablation}}
\end{figure*}

\section{Introduction to HSS and sparse-HSS}

\subsection{Introduction}
Hierarchically Low Rank Compression (such as HSS) methods exploit low-rank structure in off-diagonal blocks to enable fast matrix operations.  In a two-level HSS, a matrix $A\in\mathbb{R}^{N\times N}$ is partitioned into two diagonal blocks $A_{11},A_{22}$ and off-diagonals $A_{12},A_{21}$ that are approximated by low-rank factors.  We build upon this by introducing a three-level HSS that recursively splits each diagonal block and achieves greater compression on matrices with multiple scales of low-rank structure.

\subsection{Background: Two-Level HSS}
Given 
$A=\begin{bmatrix}A_{11} & A_{12} \\
             A_{21} & A_{22}\end{bmatrix},$
We compute low-rank SVD truncations
$A_{12}\approx U_0R_0,\quad A_{21}\approx U_1R_1,$
and store $D_0=A_{11},D_1=A_{22}$.  The mat-vec $y=A x$ then costs
$O(Nr)$ ($r$ being the r-rank factorization of $A_{12}$ and $A_{21}$) instead of $O(N^2)$, via $y_1 = D_0x_1 + U_0(R_0x_2),\quad y_2 = U_1(R_1x_1) + D_1x_2.
$

\subsection{Three-Level HSS Construction}
We further partition $D_0$ and $D_1$ into $2 \times 2$ contiguous blocks of size $N/2 \times N/2$, yielding a three-level binary tree:
\begin{align}
\underbrace{\bigl(\underbrace{(00,01)}_{D_0}\,,\underbrace{(10,11)}_{D_1}\bigr)}_{\text{root  level}}. 
\end{align}
We use the same notation as in Algorithm \ref{alg:three-level}.
We subsequently perform two-level compression at each node, that is, for $D_0$ and $D_1.$ The full algorithm and recursion are shown in Algorithm \ref{alg:three-level}. We see that at line 4, root level compression is called on weight matrix $ A$ then recursively the compression is called for left diagonal block $D_0$ and right child diagonal block $D_1$ in lines 5 and 6 respectively. In the algorithm, we also have a tolerance parameter $\epsilon$ to drop the singular values below the tolerance $\epsilon.$ In our experiments, we fix this tolerance to be $1e-6$ throughout. We also have the parameter $k,$ for rank-$k$ compression.

\subsection{HSS Matrix-Vector Multiplication}
Given the HSS generators from above, the product $y=A x$ is computed by first applying two-level mat-vec on each subtree, then adding top-level coupling. For brevity we skip the matrix vector detail.

\subsection{\label{sec:shss}Sparse plus Hierarchical Low Rank}
The sparse plus HSS is a slight but important modification to HSS in the same way as it is done for sparse plus low rank with SVD before. The only difference is that now it is done hierarchically. Also, the original rank parameter is reduced by half at each step of recursion, this is because the block dimensions reduce to half. In the following, we briefly describe the three-level sparse plus HSS. As a first step (1) we take out a sparse significant entries ($>$ tol, with some tolerance $\epsilon$ predefined as in Algorithm \ref{alg:three-level}) in matrix $S^1_0$ with the residual matrix $A = S^1_0 + A^1_0.$ (2) Then we do RCM reordering to bring the largest entries along the diagonal blocks to obtain $\widehat{A}^1_0,$ where $\widehat{A}^1_0$ is the matrix obtained after applying the RCM reordering. The RCM reordering corresponds to applying row and column permutation using permutation matrix $ p^1_0$ we save this at each level of recursion, because we need this during matrix-vector operations. (3) We then perform the {\bf first} level of hierarchical low rank compression on $\widehat{A}^1_0$ using Algorithm \ref{alg:three-level}, which gives the output $\{D_{0},D_{1},U_{t0},R_{t0},U_{t1},R_{t1} \},$ where $D_{0}$ and $D_{1}$ are (1,1) and (2,2) blocks; these are {\em unmodified} block diagonals of the matrix $\widehat{A}^1_0,$ and $\{ U_{t0},R_{t0} \}$ and $\{ U_{t1},R_{t1} \}$ are low rank factorizations of off-diagonal blocks $(1,2)$ and $(2,1)$ blocks of $\widehat{A}^1_0.$ We may visualize this as follows:
\begin{align*}
    \widehat{A}^1_0 = \begin{pmatrix}
       D_{0} & U_{t0}R^T_{t0}  \\ 
       U_{t1}R^T_{t1} & D_{1}
    \end{pmatrix},
\end{align*}
where the low rank factorizations $U_{t0}R^T_{t0}$ and $U_{t1}R^T_{t1}$ can be achieved using randomized SVD discussed before.
We essentially repeat the three steps above for $D_{0}$ and $D_{1}.$ That is, we take out the significant components of $D_0$ and $D_1$ into sparse matrices $S_0$ and $S_1$ respectively to obtain the residual matrices, which are then permuted using RCM to obtain $\widehat{D}_0$ and $\widehat{D}_1.$ We save the corresponding permutation matrices as $p^2_0$ and $p^2_1$ for $\widehat{D}_0$ and $\widehat{D}_1$ respectively. We then do {\bf second} level of compression on the blocks. The compression for the block $\widehat{D}_0$ gives $\{D_{00},D_{01},U_{00},R_{00},U_{01},R_{01} \}$ and similarly the compression of the block $\widehat{D}_1$ gives us the output $\{ D_{10},D_{11},U_{10},R_{10},U_{11},R_{11} \}$ as shown in Algorithm \ref{alg:three-level}. Here, as before $\{ D_{00},D_{01} \}$ are {\em unmodified} diagonal blocks of $\widehat{D}_0$ and $\{ D_{10},D_{11} \}$ are {\em unmodified} diagonal blocks of $\widehat{D}_1.$ Here, $\{ U_{00},R_{00} \}$ and $\{ U_{01},R_{01} \}$ are the low rank factorization of the off-diagonal blocks of $\widehat{D}_0$ and  $\{ U_{10},R_{10} \}$ and $\{ U_{11},R_{11} \}$ are the low rank factorizations of the off-diagonal blocks of $\widehat{D}_1.$ We may visualize these as follows:
\begin{align*}
    \widehat{D}_0 = \begin{pmatrix}
       D_{00} & U_{00}R^T_{00}  \\ 
       U_{01}R^T_{01} & D_{01}
    \end{pmatrix}, \quad \widehat{D}_1 = \begin{pmatrix}
       D_{10} & U_{10}R^T_{10}  \\ 
       U_{11}R^T_{11} & D_{11}
    \end{pmatrix}.
\end{align*}
We may now repeat the idea recursively to do sparse plus HSS compression of the diagonal blocks $D_{00}$ and $D_{01}.$ A housekeeping of sparse matrices at each level and the subsequent permutation matrices applied to the residual matrices is required. We note that matrix vector operations such as $y = Ax$ are required in the forward pass of LLM, and the input vector $x$ is first permuted, then the inverse permutation is applied to the output vector $y.$

\section*{Inference (Matrix-Vector Multiplication)}

As discussed in Section 4.5, the sHSS-RCM compression results in a model where each compressed weight matrix $W$ is stored as a collection of sparse matrices ($S$), permutation matrices ($P$), hierarchical low-rank factors ($U, R$), and small dense diagonal blocks ($D$) from each level of the recursion.

The inference operation $y = Wx$ cannot be performed with a single matrix multiplication. Instead, it is computed as a recursive, multi-step process that reverses the compression. This process is highly efficient as it replaces one large $O(N^2)$ operation with a series of sparse and narrow $O(Nr)$ operations (where $r$ is the small rank).

Let us consider the three-level compression described in the paper. The full matrix $W$ is approximated as:

\[
W \approx S_0^1 + 
P_0^1 
\left( 
\begin{pmatrix} 
S_0 & 0 \\ 
0 & S_1 
\end{pmatrix} 
+ 
\begin{pmatrix} 
\hat{D}_0 & U_{t0}R_{t0} \\ 
U_{t1}R_{t1} & \hat{D}_1 
\end{pmatrix} 
\right) 
(P_0^1)^T
\]

where $\hat{D}_0$ and $\hat{D}_1$ are themselves compressed recursively.

The matrix-vector product $y = Wx$ is computed as follows:

\begin{enumerate}
    \item {Top-Level Sparse Multiply:} Compute the contribution from the top-level spikes:
    \[
    y_S = S_0^1 x.
    \]

    \item {Permute Input Vector:} Shuffle the input vector using the top-level RCM permutation:
    \[
    x_{\text{shuffled}} = P_0^1 x.
    \]

    \item {Recursive HSS Multiply:}  
    Multiply $x_{\text{shuffled}}$ by the hierarchical structure.  
    Split $x_{\text{shuffled}}$ into $x_a$ and $x_b$ corresponding to blocks $\hat{D}_0$ and $\hat{D}_1$.

    \begin{itemize}
        \item {Level 2 Call (Left Block):}  
        Compute:
        \[
        y_a = \hat{D}_0 x_a,
        \]
        which recursively applies steps (1)--(4): permuting $x_a$ using $P_0^2$, multiplying by lower-level HSS factors $(D_{00}, D_{01})$, adding the sparse part $S_0$, and applying $(P_0^2)^T$.

        \item {Level 2 Call (Right Block):}  
        Similarly compute:
        \[
        y_b = \hat{D}_1 x_b.
        \]

        \item {Level 1 Combine:}  
        Combine results using the top-level HSS factors:
        \[
        y_{\text{HSS\_shuffled}} = 
        \begin{pmatrix} y_a \\ y_b \end{pmatrix}
        +
        \begin{pmatrix}
            U_{t0}(R_{t0}x_b) \\
            U_{t1}(R_{t1}x_a)
        \end{pmatrix}.
        \]
    \end{itemize}

    \item {Inverse-Permute Output:}  
    Unshuffle using the inverse RCM permutation:
    \[
    y_{\text{HSS}} = (P_0^1)^T y_{\text{HSS\_shuffled}}.
    \]

    \item {Final Result:}  
    \[
    y = y_S + y_{\text{HSS}}.
    \]
\end{enumerate}

\section{Numerical Experiments}
All our experiments were conducted on a single NVIDIA H100 GPU. Most of the compute-intensive operations were GPU-accelerated, particularly leveraging the PyTorch library for Singular Value Decomposition (SVD) and associated tensor operations. The primary model selected for compression was LLaMA 7B base \cite{touvron2023llamaopenefficientfoundation}, which contains approximately 6.7 billion parameters. Among the layers, we specifically targeted the attention projection layers, namely \texttt{q\_proj}, \texttt{k\_proj}, and \texttt{v\_proj}, for structured compression. These layers alone account for around 1.6 billion parameters, which constitute a significant fraction of the model’s total capacity.

We deliberately restricted compression to these layers due to observed instability in perplexity (PPL) scores when compressing all linear layers indiscriminately. Compressing every linear layer introduces heterogeneous matrix structures and varying sparsity patterns that complicate optimization and often degrade generation quality. Investigating compression across a broader range of layers remains a compelling direction for future research, particularly with layer-wise adaptation strategies.

Moreover, the selected projection weight matrices are inherently square, making them ideal candidates for symmetric matrix reordering techniques such as Reverse Cuthill–Mckee (RCM). These reorderings are known to reduce bandwidth and improve compressibility by concentrating non-zero elements near the diagonal. Extending such reorderings to non-square (rectangular) matrices, such as those found in MLP layers, presents additional challenges, including asymmetric padding and the construction of pseudo-permutation matrices. We encourage interested readers to explore this frontier, particularly in adapting classical graph-based reordering techniques to the rectangular setting.

We also emphasize that all experiments were performed using 16-bit floating point precision (fp16), which provides a favourable trade-off between compute efficiency and numerical stability. Although 8-bit quantization was considered, early tests showed degraded generation quality and inconsistent decoding behaviour. Thus, we avoided 8-bit compression in this study, leaving it to future work involving quantization-aware compression pipelines.

\subsection{Evaluation Dataset}

For evaluating the compressed versions of the LLaMA 7B model we turned to the WikiText-103-v1 test dataset. The WikiText-103-v1 dataset is essentially a massive language model benchmark with over 1.8 million training examples and 4,358 test examples that come from Wikipedia articles that are classified as ``Good" or ``Featured". The unique feature of this dataset is that each sample is made up of continuous, natural English text that includes original case, punctuation and numbers. We use this to test the ability of language models to generate and predict coherent, relevant text captured by perplexity score \cite{merity2016pointersentinelmixturemodels}.

\subsection{Storage versus Accuracy}
\begin{figure}[t!]
    \centering
    \includegraphics[width=0.75\linewidth]{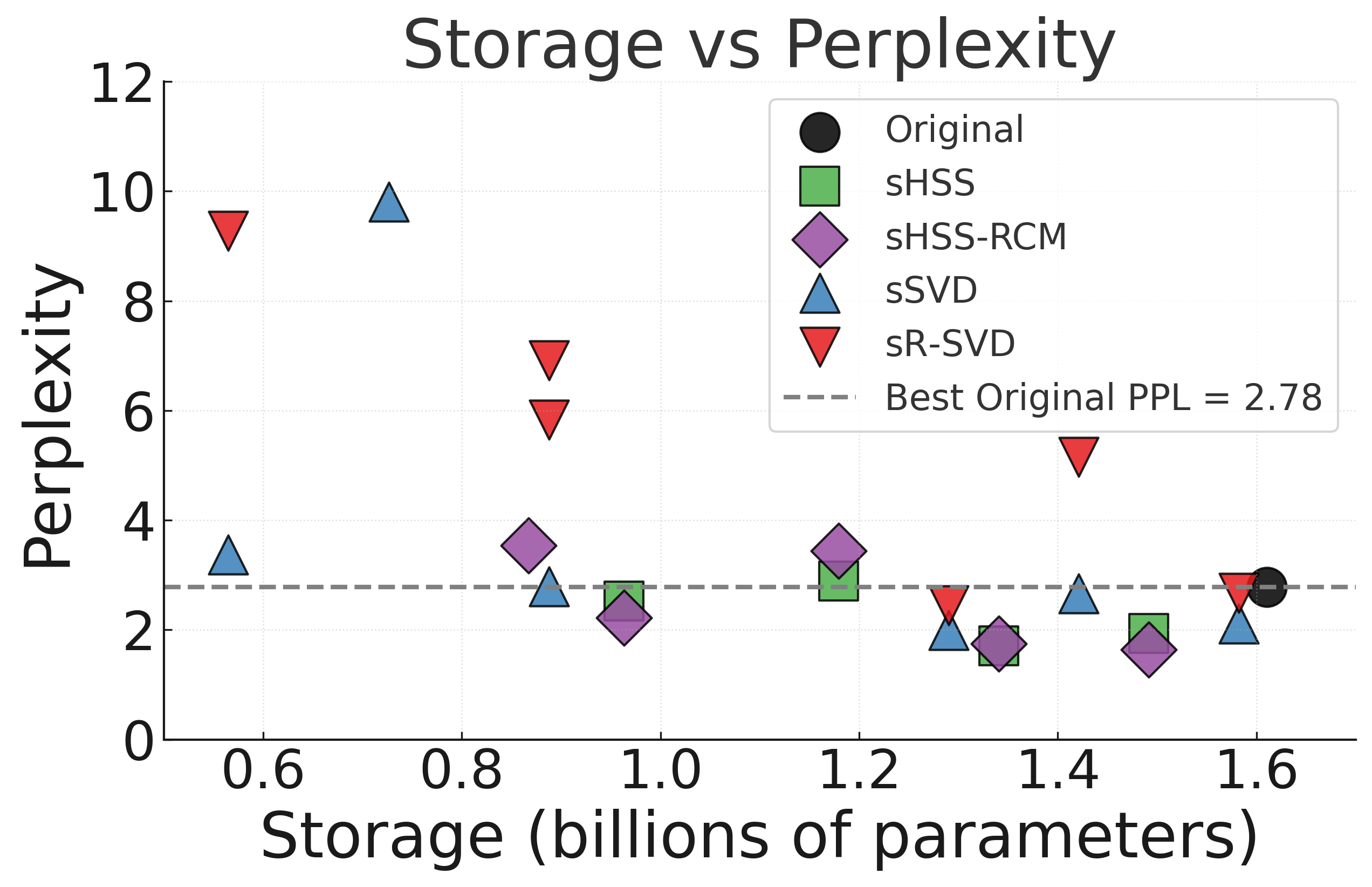}
    \caption{Storage vs accuracy plot. In this plot, we observe that in general, hierarchical methods like sHSS and RCM ordered methods like sHSS-RCM have better perplexity for a given similar range of storage requirements compared to sparse plus low rank using SVD, and for the randomised variant sR-SVD. Note that it is not possible to have comparisons with the same storage requirements. We also observe that often RCM reordering, denoted by the purple shaded diamond, has lower perplexity scores compared to the one without RCM, the green shaded square boxes.}
    \label{fig:storage-accuracy}
\end{figure}

In Figure \ref{fig:storage-accuracy}, we show storage versus perplexity scores for Llama 7b base model \cite{touvron2023llamaopenefficientfoundation}; the base uncompressed model is denoted by Original, followed by sHSS, the sparse hierarchical low rank method, then sHSS-RCM, sHSS with RCM ordering, followed by sSVD, the sparse SVD and sR-SVD, the sparse random SVD. We also show the PPL score for the Original.  We note here that storage is affected by rank and sparsity for sparse methods. Hence, larger storage is typically contributed by both. For aggressive compression, by more than 50\% of the target parameters, we observe that between 0.8B and 1.0B parameters, sHSS and sHSS-RCM do better than sRSVD and marginally better than sSVD.  For moderate compression of about 10\% of the target parameters, we find again that PPL scores of hierarchical methods have lower PPL scores than SVD-based methods. In fact, we get on average better PPL scores than the base model despite 10\% compression of parameters. We also observe that RCM ordering of the weights helps most of the time in leading to better PPL scores. For HSS, RCM is useful because it brings the max entries of entries closer to the diagonal (after removing the entries into $S$ see before). We discuss the effect of RCM reordering below.

\subsection{Role of Sparsity and Rank on PPL Scores}

The compression of the models is controlled by the sparsity parameter and the rank parameter. For a fixed outer rank, we show that cutting out a larger percentage of the entries of the current layer’s weights, denoted as “sp”, has a dramatic effect on the low-rank factorization of our SVD method. For instance, “sp10” means we’re removing 10\% of the current layer’s weights before we apply low-rank compression. The fewer the entries (when sp parameter is high) we use for low rank decomposition, the more accurate our low rank approximation becomes, which is why for a fixed rank, higher sp values lead to better PPL scores. Interestingly, we also see that the two algorithms proposed by us, sHSS and sHSS-RCM, perform the best.

\subsection{Role of RCM Reordering}
We wonder whether RCM reordering has any significant role to play in the quality of the compressed models. From Figure \ref{fig: ablation}, we find that when RCM is applied after 10\% of the matrix entries are taken away, then RCM has better PPL scores; otherwise, for the remaining two cases, we don't see much difference. However, in another experiment in Figure \ref{fig:storage-accuracy}, where we look at storage-PPL tradeoff, we see that sHSS-RCM has slightly better PPL than sHSS for the same storage; that is, we see that purple shaded diamond of sHSS-RCM is just below the green-shaded square of sHSS. In general, from our experiments, we conclude that it gives a slight gain with RCM reordering. However, more experiments on larger LLM models may be required to conclude on the significant benefits of RCM.

\section*{Conclusion}
We introduced a hierarchical sparse–plus–low-rank framework that unifies pruning, SVD and HSS compression under a single algebraic lens and introduced two hierarchical compression methods \textit{sHSS} and \textit{sHSS-RCM}.
By explicitly separating significant absolute entries from weight matrices, followed by low rank compression, our method attains a stronger trade-offs between perplexity and more optimal storage  than state-of-the-art sparse–SVD baselines for compression of layers of Llama-7B.
Reverse Cuthill–Mckee permutations further reduce off-diagonal ranks, demonstrating that RCM re-ordering can be a useful preprocessing step in linear layer compression for LLM.
Extensive ablations across sparsity ratios (10–30\%), HSS depths and rank budgets confirm the robustness of the approach, while a GPU-friendly implementation keeps compression time within minutes on a single H100.
Importantly, the compressed models retain full FP16 inference speed because the sparse mask and low-rank factors are applied with batched CUDA kernels.
Our implementation is in pytorch, and can be extended to compression of larger LLM paving the way for practical billion-parameter LLM deployment on commodity hardware.
The empirical gains suggest that hierarchical structure in weight matrices is under-exploited and complementary to quantization or knowledge-distillation pipelines.
Looking forward, combining sHSS with 4-bit weight quantization and dynamic rank adaptation during fine-tuning are promising avenue to push compression beyond the 7-billion-parameter frontier.
We also plan to extend the technique to transformer activations, enabling end-to-end memory reduction during training.
Overall, hierarchical sparse plus low-rank compression offers a scalable path toward compact foundation models. We explored RCM reordering, but there are quite a few sparse reordering techniques popular in scientific computing that could be explored such as Nested Dissection \cite{kumar2013b,kumar2010b,kumar2014c}, AMD \cite{Amestoy2004}, and COLAMD \cite{davis2004}.

\section*{LLM Usage}
ChatGPT-5.1 free version was used to polishing only the text, and rephrasing most paragraphs.

\bibliographystyle{ACM-Reference-Format}

\bibliography{custom}

\appendix

\end{document}